# A Computer Vision System for Attention Mapping in SLAM based 3D Models

Lucas Paletta, Katrin Santner, Gerald Fritz, Albert Hofmann,
Gerald Lodron, Georg Thallinger, Heinz Mayer

JOANNEUM RESEARCH Forschungsgesellschaft mbH,
DIGITAL - Institute for Information and Communication Technologies
Steyrergasse 17, 8010 Graz, Austria

*{lucas.paletta, katrin.santner, gerald.fritz, albert.hofmann, gerald.lodron,
georg.thallinger, heinz.mayer}@joanneum.at*

**Abstract**

*The study of human factors in the frame of interaction studies has been relevant for usability engineering and ergonomics for decades. Today, with the advent of wearable eye-tracking and Google glasses, monitoring of human factors will soon become ubiquitous. This work describes a computer vision system that enables pervasive mapping and monitoring of human attention. The key contribution is that our methodology enables full 3D recovery of the gaze pointer, human view frustum and associated human centred measurements directly into an automatically computed 3D model in real-time. We apply RGB-D SLAM and descriptor matching methodologies for the 3D modelling, localization and fully automated annotation of ROIs (regions of interest) within the acquired 3D model. This innovative methodology will open new avenues for attention studies in real world environments, bringing new potential into automated processing for human factors technologies.*

## 1. Introduction

The study of human factors in the frame of interaction studies has relevant in usability engineering and ergonomics for decades [18]. Today, with the advent of wearable eye-tracking and Google glasses, monitoring of human visual attention and the measuring of human factors will soon become ubiquitous. This work describes a computer vision system that enables pervasive mapping and monitoring of human attention. The key contribution is that our methodology enables full 3D recovery of the gaze pointer, human view frustum and correlated human centred measurements directly into an automatically computed 3D model, in real-time. It applies RGB-D SLAM and descriptor matching methodologies for the 3D modelling, localization and fully automated annotation of ROIs (regions of interest) within the acquired 3D model.

This work presents a computer vision system methodology that, firstly, enables to precisely estimate the position and orientation of human view frustum and gaze and from this enables to precisely analyse human attention in the context of the semantics of the local environment (objects [5], signs, scenes, etc.). Figure 1 visualizes how accurately human gaze is mapped into the 3D model for further analysis. Secondly, the work describes how ROIs are automatically mapped from a reference video into the model and from this prevents from state-of-the-art laborious manual labelling of tens / hundreds of hours of eye tracking video data. This will provide a scaling up of nowadays still small sketched usability studies with ca. 10-15 users and thus enable for the first time statistically significant usability engineering studies.

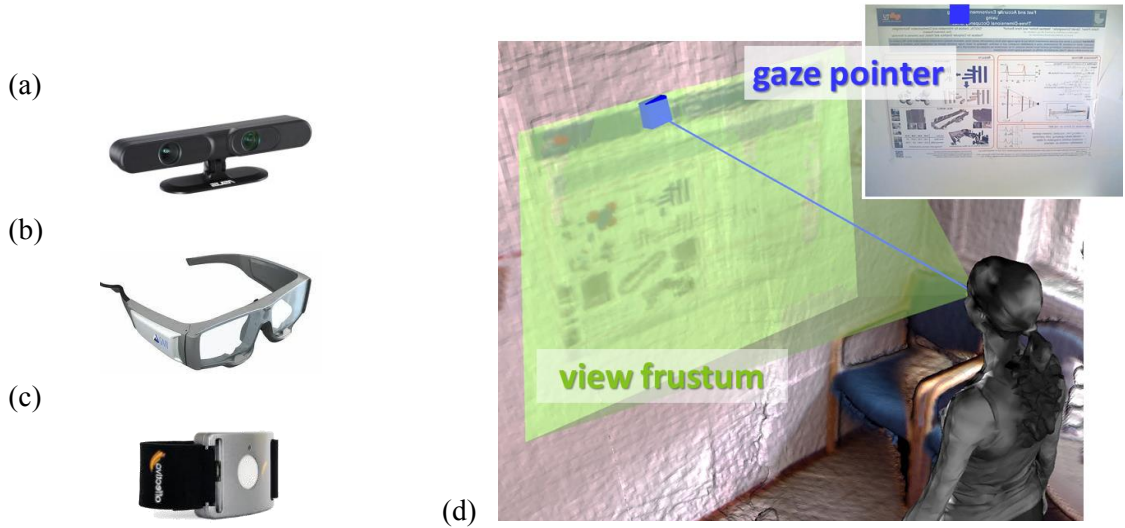

Figure 1. *Sketch of sensors used in the study (left) and typical gaze recovery (d), right. A full 6D recovery of the view frustum and gaze (right) is continuously mapped into the 3D model. (a) RGB-D scanning device, (b) eye tracking glasses (ETG) and (c) bio-electrical signal device.*

The methodology for the recovery of human attention in 3D environments is based on the workflow as sketched in Figure 2: For a spatio-temporal analysis of human attention in the 3D environment, we firstly build a spatial reference in terms of a three-dimensional model of the environment using RGB-D SLAM methodology [15]. Secondly, the user's view is gathered with eye tracking glasses (ETG) within the environment and localized from extracted local descriptors [10,13]. Then ROIs are marked on imagery and automatically detected in video [10] and then mapped into the 3D model. Finally, the distribution of saliency onto the 3D environment is computed for further human attention analysis, such as, evaluation of the attention mapping with respect to object and scene awareness. Saliency information can be aggregated and, for example, being further evaluated in the frame of user behaviours of interest. The performance evaluation of the presented methodology firstly refers to results from a dedicated test environment [14], where we demonstrate very low angular and Euclidean projection errors (≈0.6º, ≈1.1 cm) on average which therefore enables to capture attention on daily objects and activities (package logos, cups, books, pencils).

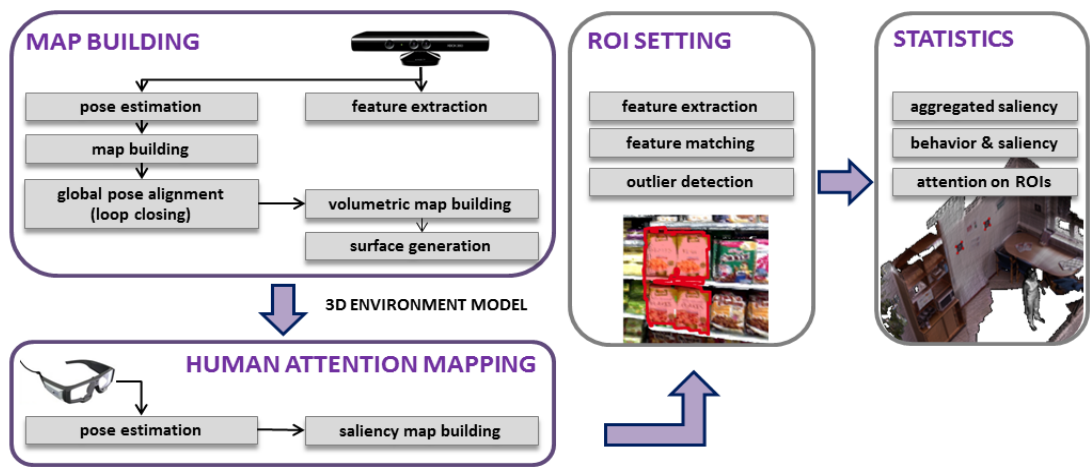

Figure 2. *Sketch of workflow for 3D gaze recovery and semantic ROI analytics*

## 2. Related Work

**Human Attention Analysis in 3D**. 3D information recovery of human gaze has in principle been targeted before. Munn et al. [12] introduced monocular eye-tracking and triangulation of 2D gaze positions of subsequent key video frames, obtaining observer position and gaze pointer in 3D. However, they reconstructed only single 3D points without reference to a 3D model with angular errors of ≈3.8° (compared to our ≈0.6°). Voßkühler et al. [20] analyzed 3D gaze movements with the use of a special head tracking unit, necessary for their intersection of the gaze ray with a digitized model of the surrounding. Pirri et al. [17] used for this purpose a mass marketed stereo rig that is required in addition to a commercial eye-tracking device, and attention cannot be mapped and tracked. The achieved accuracy indoor is ≈3.6 cm at 2 m distance to the target compared to our ≈0.9 cm [14]. In general, we present a straight forward solution of mapping fixation distributions onto a 3D model of the environment; the presented work extends through the automated annotation process.

**Vision Based Dense Reconstruction**. Vision based Simultaneous Localization and Mapping (SLAM) aims at building a map of a previously unknown environment while simultaneously estimating the sensors pose within this map. In the last years SLAM has been performed using a huge variety of visual sensor such as single cameras [1], stereo or trinocular systems. With the launch of range image devices, large scale dense reconstruction of indoor environments has been proposed, such as, Newcombe et al. [7], presenting their real-time dense tracking and mapping system of small desktop scenes named KinectFusion. Dense reconstruction of large cyclic indoor environments has been presented via bundle adjustment techniques and fusion of probabilistic occupancy grid maps with loop closing [16].

**Vision Based Localization and Logo Detection**. Recently, several authors proposed a least-squares optimization routine minimizing the re-projection error, others perform a perspective n-Point pose estimation algorithm. Both groups rely on correspondences established between 3D model points and 2D image points. In case of a large scale map consisting of thousands of model points, correspondence estimation becomes computationally too expensive. Therefore, image retrieval techniques [13] have been proposed to reduce the number of possible matching candidates. Logo detection is done on more general reference material in our work. The base for reference logos is packaging similar to those contained in the Surrey Object Image Library. An evaluation of state-of-the-art algorithms in machine vision object recognition on image databases shows that SIFT [10] performs best on comparative image databases (SOIL-47 dataset [2]).

## 3. Gaze Localization in 3D Models

**Visual Map Building and Camera Pose Estimation**. For realistic environment modeling we make use of an RGB-D sensor (e.g. ASUS Xtion ProLive[1]) providing per pixel color and depth information at high frame rates. After intrinsic and extrinsic camera calibration, each RGB pixel can be assigned a depth value. Since we are interested in constructing a 3D environment in reasonable time, we perform feature based visual SLAM relying on the approach of [15]. Our environment consists of a sparse point cloud, where each landmark is attached a SIFT [10] descriptor for data association during pose tracking and for vision based localization of any visual device within this reconstructed environment. Estimated camera poses (key frames) are stored in a 6DOF manner. Incremental camera pose tracking, assuming an already existing map, is done by key point matching followed by a least-square optimization routine minimizing the re-projection error of 2D-3D correspondences. New landmarks are established using the previously estimated camera pose and the depth information stemming from the RGB-D device. Finally, sliding-window bundle adjustment is performed to refine both camera and landmark estimates. To detect loop closures we use a bag-of-words approach [13].

---

[1] http://www.asus.com/Multimedia/Xtion_PRO_LIVE/

**Densely Textured Surface Generation**. For realistic environment visualization, user interaction and subsequent human attention analysis, a dense, textured model of the environment has to be constructed. Therefore, depth images are integrated into a 3D occupancy grid [15] using the previously corrected camera pose estimates. Hereby, we follow the pyramidal mapping approach presented in [16] implemented on the GPU. The whole volume is divided into sub-volumina, whose sizes depend on the memory architecture of the GPU (typically $512^3$ voxels). Unused sub-volumina (e.g. already mapped or not visible) are cached in CPU memory (or any arbitrary storage devices) and reloaded on demand. Realistic surface construction is done by a marching cubes algorithm [9], where overlapping sub-volumina guarantee a watertight surface. To apply realistic texture, we use a simple per vertex coloring approach. Each vertex' RGB color value is computed by projecting it onto the color image plane and taking the running average over all possible values resulting in a smooth, colored mesh (see Figure 3).

**3D Gaze Recovery from Monocular Localization.** To estimate the proband's pose, SIFT key points are extracted from ETG video frames and then matched landmarks from the prebuilt environment and a *full 6DOF pose* is estimated using the perspective n-Point algorithm [8]. Given the proband's pose together with the image gaze position, we are interested in its fixation point within the 3D map. Therefore, we compute the intersection of the viewing ray through the gaze position with the triangle mesh of the model. For rapid interference detection we make use of an object oriented bounding box tree [4] reporting the surface triangle and penetration point hit by the ray. Fixation hits are integrated over time resulting in a *saliency map* used to study and visualize each user's attention in the 3D environment (see Figure 4), and finally we apply a Gaussian weighting kernel over nearby surface triangles.

**Automated 3D Annotation of Regions of Interest.** Annotation of ROIs in 2D or even 3D information usually causes a process of massive manual interaction. In order to map objects of interests, such as, logos, package covers, etc. into the 3D model, we first use logo detection in the high resolution scanning video to search for occurrences of predefined reference appearances. We apply the SIFT descriptor [10] to find the appropriate logo in each input frame. Visual tracking has been omitted so far in order not to introduce tracking errors into the 3D mapping step (see Figure 5). For robustness, ROI polygons are filtered if the geometric transformation of nearby frames significantly differs from the identity transform. For ROI identification in 3D model space, we use the key frame poses estimated as described above together with the ROIs automatically detected in the associated image. Each surface is projected onto the image plane and checked for being inside the ROI polygon. Resulting ROIs in the image and 3D domain are depicted in Fig.5.

## 4. Experimental Results

**Eye Tracking Device.** The mass marketed SMI™ eye-tracking glasses (Figure 1b) - a non-invasive video based binocular eye tracker with automatic parallax compensation - measures the gaze pointer for both eyes with 30 Hz. The gaze pointer accuracy of 0.5°–1.0° and a tracking range of 80°/60° horizontal/vertical assure a precise localization of the human's gaze in the HD 1280x960 scene video with 24fps. An accurate three point calibration (less than 0.5° validation error) was performed and the gaze positions within the HD scene video frames were used for further processing. To evaluate our system in a realistic environment we recorded data on a shop floor covering an area of about 8x20m². We captured 2366 RGB-D images and reconstructed the environment consisting of 41700 natural visual landmarks and 608 key frames. The results are shown in Figure 3c.

**Recovery of 3D gaze.** In the study on human attention, 3 proband's were wearing eye-tracking glasses and the Affectiva Q sensor for measuring electrodermal activity (EDA) and accelerometer data. Probands had the task to search for three specific products which define the ROI in 2D and 3D. The ratio of successfully localized versus acquired frames is described in in Table 1: user 1 and 2 are efficiently localized. In general, blurred imagery, images depicting less modeled area in the test environment and too close views of the scenery cause localization outages that will be improved with appropriate tracking methodology as future work. However, the system allows a fully automatic computation of each proband's path within the environment, the full recovery of gaze and aggregation of saliency over time within the 3D model (Figure 3). The accuracy estimation of the proposed 3D gaze recovery has been reported in [14] with an angular projection error of ≈0.6º within the chosen 3D model which is therefore smaller than the calibration error of the eye-tracking glasses (≈1º). The Euclidean projection error was only ≈1.1 cm on average and thus enables to capture attention on daily objects and activities (packages, cups, books, pencils).

**ROI detection in 2D and 3D.** To evaluate the performance of automated 3D ROI association from HD video, we generated ground truth data, where each occurrence is annotated as shown in Figure 5. For accuracy evaluation of the ROI detection algorithm, we are only interested in the detection performance, since tracking is not relevant for the 3D mapping and robust detections are preferred instead of continuous tracks containing imprecise regions. Since a ROI may be composed of multiple parts, the *temporal coverage* by detections of each gives an indication whether the complete ROI is covered. We have three ROIs consisting of eight parts with a temporal coverage between 37% and 98%. The precision and recall values act as a common quality measure of detection performance. Here, we employ the following overlap criterion $O(R)$ of a ROI $R$ and its ground truth $G(R)$. The results are presented in Table 2. While having high precision rates at ROI #1 and #3, very similar breakfast cereal products causes the algorithm to produce false positives resulting in a low precision rate for ROI #2 (see Figure 5). Given the high recall and spatial overlap values, the logo detection algorithm provides suitable input for the 3D mapping procedure.

To evaluate the reliability, correctness and accuracy of the automatic ROI computation in the three-dimensional domain, we manually segmented the appropriate ROIs in 3D space as ground truth. As an accuracy measure we again employ the overlap criterion defined above. To show the influence of the accuracy of the ROI detection in the image domain, we compare the three-dimensional ROIs computed out of the *fully automatic* two-dimensional ROI detection algorithm against the *manually annotated* ones. The results are given in Table 3. Clearly, false, missing or imprecise detections in the image domain produce a high error in the 3D space, since the overlap criterion for manually annotated ROIs is higher in each case. This is also visualized in Figure 5, where a single 2D detection outlier results in erroneous 3D ROIs.

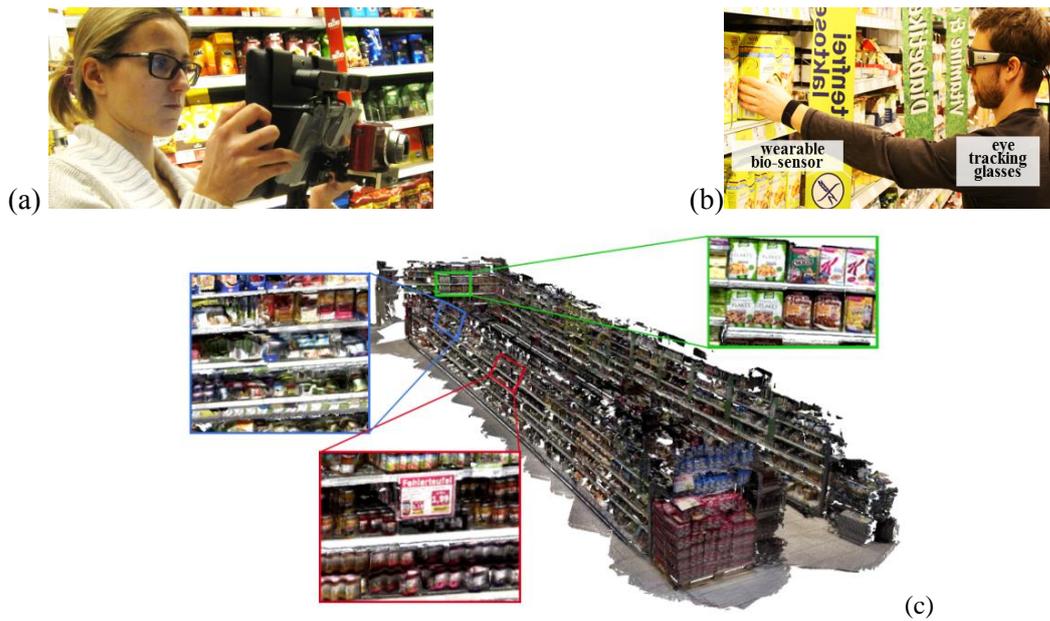

**Figure 3.** Hardware (a) for the 3D model building process (Kinect and HD camera), (b) study with packages, (c) 3D model of the study environment, a shop floor for experimental studies.

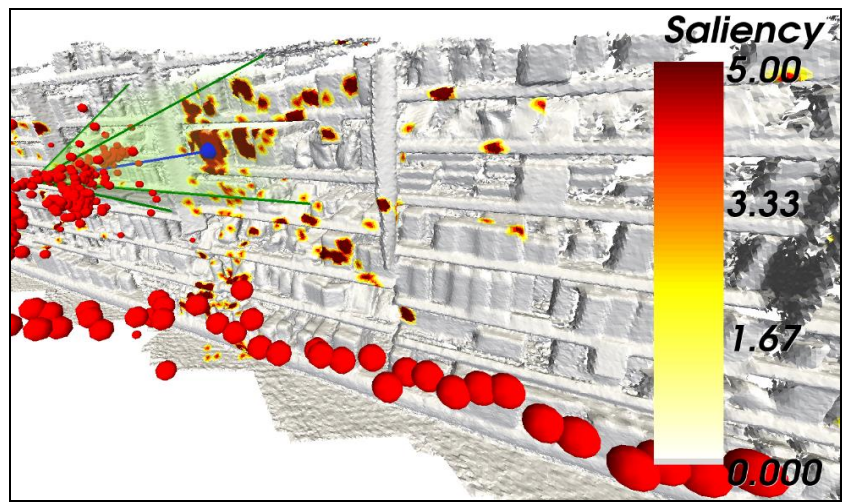

*Figure 4. Mapping of saliency onto the acquired 3D model and automated recovery of the ETG position trajectory.*

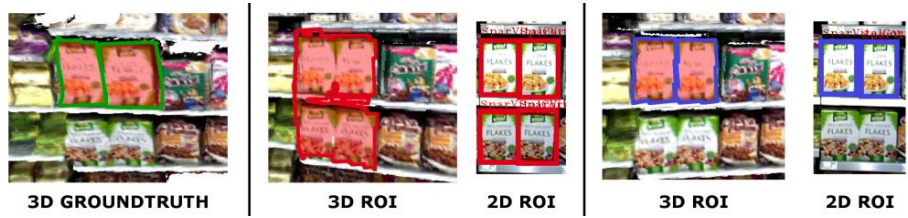

**Figure 5.** *Automated ROI detection in the 2D and 3D domain. (left) 3D ground truth annotation. (mid) ROIs computed in 3D out of automatic 2D detections. (right) 3D ROIs computed out of the 2D ground truth data.*

**Table 1.** *Performance of localization in two typical user tracks.*

| proband | # frames in total | # frames successfully localised |
|---|---|---|
| 1 | 1903 | 1512 (79.45%) |
| 2 | 1306 | 1088 (83.31%) |

Table 2. *Accuracy evaluation of the ROI detection algorithm.*

| performance \ ROI | # 1 | # 2 | # 3 | total |
|---|---|---|---|---|
| # ground truth annot. | 21 | 87 | 95 | 203 |
| # logo detections | 19 | 184 | 82 | 285 |
| # true positives | 19 | 86 | 70 | 175 |
| precision | 1.00 | 0.47 | 0.85 | 0.61 |
| recall | 0.90 | 0.99 | 0.74 | 0.86 |
| avrg. spatial overlap ($\sigma$) | 0.87 (0.02) | 0.90 (0.06) | 0.86 (0.05) | 0.88 (0.06) |

Table 3. *Evaluation results of the three-dimensional ROI computation.*

| R | O( R ) 2D automatic ROI detection | O( R ) 2D manual ROI annotation |
|---|---|---|
| # 1 | 0.728369 | 0.731168 |
| # 2 | 0.327668 | 0.848437 |
| # 3 | 0.611904 | 0.671895 |

**Semantic Mapping of Attention.** The proposed system allows a fully automated workflow to evaluate human attention performance entirely within a 3D model. The automatic detection of ROIs in three-dimensional space enables the system to provide the user with statistical evaluation without any manual annotation, which is known to be time consuming and error prone.

One of the basic indicators when dealing with ROIs is called AOI hit, which states for a raw sample or a fixation that its coordinate value is inside the ROI [2]. ROI #1 received the maximum hits (287) by all users, with the maximum hits counted for user #1 (112 hits). Another example is the dwell – often known as 'glance' in human factor analysis – and defined as one visit in an ROI, from entry to exit. The maximum mean dwell time was measured for ROI #1 but by user #3 (133.3 ms). Figure 6 plots the distribution of the dwell times for ROI #1 over all participants. Notice that dwells shorter than 35ms are excluded from the plot to enhance the readability. There are in total 287 hits for ROI #1, 45 visits of the region took at least 35 ms and the longest visit lasted 733.3 ms. From these data we conclude that only a minority of the captured fixations is related to human object recognition since this is known to trigger from 100 ms of observation / fixation. However, the investigation of human attention behavior is dedicated to future work and we believe that we developed, a most promising technology, in particular, for the purpose of studying mobile eye tracking in the field, in the real world, and for computational modeling of attention modeling.

## 5. Conclusions

We presented a complete system for (i) wearable data capturing, (ii) automated 3D modeling, (iii) automated recovery of human pose and gaze, and (iv) automated ROI-based semantic interpretation of human attention, using mass marketed equipment. The examples from a first relevant user study demonstrate the potential of our computer vision system to perform automated analysis and/or evaluation of the human factors, such as attention, using the acquired 3D model as a reference frame for gaze and semantic mapping, and with satisfying accuracy in the mapping from eye tracking glasses based video onto the automatically acquired 3D model. The presented system represents a significant first step towards an ever improving mapping framework for quantitative analysis of human factors in environments that are natural in the frame of investigated tasks. Future work will focus on improved tracking of the human pose across image blur and uncharted areas as well as on studying human factors in the frame of stress and emotion in the context of the 3D space.

## 6. Acknowledgments

This work has been partly funded by the European Community's Seventh Frame-work Programme (FP7/2007-2013), grant agreement n°288587 MASELTOV, and by the Austrian FFG, contract n°832045, Research Studio Austria FACTS.